\documentclass{article}

\usepackage{nips_2016}

\usepackage[utf8]{inputenc} 
\usepackage[T1]{fontenc}    
\usepackage{hyperref}       
\usepackage{url}            
\usepackage{booktabs}       
\usepackage{amsfonts}       
\usepackage{nicefrac}       
\usepackage{microtype}      
\usepackage{graphicx}
\usepackage{tabularx}
\usepackage{xcolor}

\definecolor{light-gray}{gray}{0.95}
\newcommand{\code}[1]{\colorbox{light-gray}{\texttt{#1}}}

\title{pomegranate: fast and flexible probabilistic modeling in python}

\author{
  Jacob Schreiber \\
  Paul G. Allen School of Computer Science \\
  University of Washington \\
  Seattle, WA 98195\\
  \texttt{jmschr@cs.washington.edu} \\
}

\begin{document} 
\maketitle

\begin{abstract}
We present pomegranate, an open source machine learning package for probabilistic modeling in Python. Probabilistic modeling encompasses a wide range of methods that explicitly describe uncertainty using probability distributions. Three widely used probabilistic models implemented in pomegranate are general mixture models, hidden Markov models, and Bayesian networks. A primary focus of pomegranate is to abstract away the complexities of training models from their definition. This allows users to focus on specifying the correct model for their application instead of being limited by their understanding of the underlying algorithms. An aspect of this focus involves the collection of additive sufficient statistics from data sets as a strategy for training models. This approach trivially enables many useful learning strategies, such as out-of-core learning, minibatch learning, and semi-supervised learning, without requiring the user to consider how to partition data or modify the algorithms to handle these tasks themselves. pomegranate is written in Cython to speed up calculations and releases the global interpreter lock to allow for built-in multithreaded parallelism, making it competitive with---or outperform---other implementations of similar algorithms. This paper presents an overview of the design choices in pomegranate, and how they have enabled complex features to be supported by simple code. The code is available at \url{https://github.com/jmschrei/pomegranate}
\end{abstract}


\section{Introduction}
The Python ecosystem is becoming increasingly popular for the processing and analysis of data. This popularity is in part due to easy-to-use libraries such as numpy \citep{numpy}, scipy \citep{scipy}, and matplotlib \citep{matplotlib} that aim to provide fast general purpose functionality. However, equally important are the libraries that are built on top of these to provide higher level functionality, such as pandas \citep{pandas} for data analysis, scikit-image \citep{scikit-image} for computer vision, Theano \citep{theano} for efficient evaluation of mathematical expressions, gensim \citep{gensim} for topic modeling in natural language processing, and countless others. Naturally, many machine learning packages have also been developed for Python, including those that implement classic machine learning algorithms, such as scikit-learn \citep{scikit-learn}, mlpy \citep{mlpy}, shogun \citep{shogun}, and xgboost \citep{xgboost}.

pomegranate fills a gap in the Python ecosystem that encompasses building probabilistic machine learning models that utilize maximum likelihood estimates for parameter updates. There are several packages that implement certain probabilistic models in this style individually, such as hmmlearn for hidden Markov models, libpgm for Bayesian networks, and scikit-learn for Gaussian mixture models and naive Bayes models. However, pomegranate implements a wider range of probabilistic models and does so in a more modular fashion than these other packages, having two main effects. The first is that the addition of a new probability distribution in pomegranate allows for all models to be built using that distribution immediately. The second is that improvements to one aspect of pomegranate immediately propagate to all models that would use that aspect. For example, when GPU support was added to multivariate Gaussian distributions, this immediately meant that all models with multivariate Gaussian emissions could be GPU accelerated without any additional code. pomegranate currently includes a library of basic probability distributions, naive Bayes classifiers, Bayes classifiers, general mixture models, hidden Markov models, Bayesian networks, Markov chains, as well as implementations of factor graphs and k-means++/|| that can be used individually but primarily serve as helpers to the primary models.

There are several already existing Python libraries that implement Bayesian methods for probabilistic modeling. These include, but are not limited to, PyMC3 \citep{pymc3}, PyStan \citep{pystan}, Edward \citep{edward}, pyro \citep{pyro}, and emcee \citep{emcee}. Bayesian approaches typically represent each model parameter as its own probability distribution, inherently capturing the uncertainty in that parameter, whereas maximum likelihood approaches typically represent each model parameter as a single value. An example of this distinction is that a mixture model can either be represented as a set of probability distributions and a vector of prior probabilities, or as a set of probability distributions that themselves have probability distributions over their respective parameters (such as the mean and standard deviation, should these distributions be normal distributions) and as a dirichlet distribution representing the prior probabilities. The first representation typically specifies models that are faster to both train and perform inference with, while the second is illustrative of the type of models one could build with packages that implement Bayesian methods, such as PyMC3. Both representations have strengths and weaknesses, but pomegranate implements models falling solely in the first representation.

pomegranate was designed to be easy to use while not sacrificing on computational efficiency. Models can either be specified by writing out each of the components individually if known beforehand, or learned directly from data if not. Key features, such as out-of-core learning and parallelization, can be toggled for each model independently of the definition or method calls, typically by simply passing in an optional parameter. The core computational bottlenecks are written in Cython and release the global interpreter lock (GIL), enabling multi-threaded parallelism that typically Python modules cannot take advantage of. Lastly, linear algebra operations such as matrix-matrix multiplications are implemented using BLAS with the ability to toggle a GPU if present.

All comparisons were run on a computational server with 24 Intel Xeon CPU E5-2650 cores with a clock speed of 2.2 GHz, a Tesla K40c GPU, and 256 GB of RAM running CentOS 6.9. The software used was pomegranate v0.8.1 and scikit-learn v0.19.0. pomegranate can be installed using \code{pip install pomegranate} or \code{conda install pomegranate} on all platforms. Pre-built wheels are available for Windows builds, removing the sometimes difficult requirement of a working compiler.

\section{The API}
pomegranate provides a simple and consistent API for all implemented models that mirrors the scikit-learn API as closely as possible. The most important methods are \code{fit}, \code{from\_samples}, \code{predict} and \code{probability}. The \code{fit} method will use the given data and optional weights to update the parameters of an already initialized model, using either maximum-likelihood estimates (MLE) or expectation-maximization (EM) as appropriate. In contrast, the \code{from\_samples} method will create a model directly from data in a manner similar to scikit-learn's \code{fit} method. For simple models like single distributions this corresponds only to MLE on the input data, but for most other models this corresponds to an initialization step plus a call to \code{fit}. This initialization can range from using k-means for mixture models to structure learning for Bayesian networks. The \code{predict} method returns the posterior estimate $argmax_{M} P(M|D)$, identifying the most likely component of the model for each sample. The \code{probability} method returns the likelihood of the data given the model $P(D|M)$. The other methods include \code{predict\_proba} which returns the probability of each component for each sample $P(M|D)$, \code{predict\_log\_proba} which returns the log of the previous value, and \code{summarize} and \code{from\_summaries} that jointly implement the learning strategies detailed below.

\section{Key Features}

pomegranate supports many learning strategies that can be employed during training, including out-of-core learning for massive datasets, semi-supervised learning for datasets with a mixture of labeled and unlabeled data, and minibatch learning. In addition, one can employ multithreaded parallelism or a GPU for data-parallel speedups. These features are made possible by separating out the collection of sufficient statistics from a data set (using the \code{summarize} method) from the actual parameter update step (using the \code{from\_summaries} method).

Sufficient statistics are the smallest set of numbers needed to calculate some statistic on a dataset. As an example, fitting a normal distribution to data involves the calculation of the mean and the variance. The sufficient statistics for the mean and the variance are the sum of the weights of the points seen so far $\left( \sum\limits_{i=1}^{n}w_{i} \right)$, the sum of the weighted samples $\left( \sum\limits_{i=1}^{n}w_{i}X_{i} \right)$, and the sum of the weighted samples squared $\left( \sum\limits_{i=1}^{n}w_{i}X_{i}^{2} \right)$. The mean and variance can then be directly calculated from these three numbers using the following two equations:

\begin{equation}
\mu = \frac{\sum\limits_{i=1}^{n}w_{i}X_{i}}{\sum\limits_{i=1}^{n} w_{i}} \quad\quad\quad \sigma^{2} = \frac{\sum\limits_{i=1}^{n}w_{i}X_{i}^{2}}{\sum\limits_{i=1}^{n}w_{i}} - \left( \frac{\sum\limits_{i=1}^{n}w_{i}X_{i}}{\sum\limits_{i=1}^{n}w_{i}} \right)^{2}
\end{equation}

\vspace{1em}\noindent\textbf{Out-of-core Learning:} The additive nature of the sufficient statistics means that if one were to summarize two batches of data successively and then add the sufficient statistics together, they would get the same sufficient statistics as if they were calculated from the full data set. This presents an intuitive way to handle data sets that are too large to fit in memory, by chunking the dataset into batches that do fit in memory and summarizing them successively, adding the calculated sufficient statistics together afterwards. This can be done by passing in a \code{batch\_size} parameter to your training method, for example \code{model.fit(X, batch\_size=10000)} would train a pre-initialized model on more data than can fit in memory by successively summarizing batches of size 10,000 until the full data set has been seen. The \code{summarize} and \code{from\_summaries} methods can also be used independently to implement custom out-of-core strategies.  

\vspace{1em}\noindent\textbf{Minibatch Learning:} A natural extension of the out-of-core strategy is minibatch learning, where a parameter update is done after one or a few batches, instead of the full data set. This is in contrast to batch methods that calculate an update using the entire dataset, and stochastic methods that typically update using only a single sample. Minibatching can be specified by passing values to both \code{batch\_size} and \code{batches\_per\_epoch} parameters when using \code{fit} or \code{from\_summaries}, where the \code{batches\_per\_epoch} is the number of batches to consider before making an update.

\vspace{1em}\noindent\textbf{Semi-supervised Learning:} Semi-supervised learning is the task of fitting a model to a mixture of both labeled and unlabeled data. Typically this arises in situations where labeled data is sparse, but unlabeled data is plentiful, and one would like to make use of both to learn an informed model. pomegranate supports semi-supervised learning for \code{HiddenMarkovModel}, \code{BayesClassifier}, and \code{NaiveBayes} models as a combination of EM and MLE. Models are initialized using MLE on the labeled data. Next, a version of EM is used that combines the sufficient statistics calculated from the labeled data using MLE with the sufficient statistics calculated from the unlabeled data using EM at each iteration until convergence. This is automatically toggled whenever -1 is present in the label set, following scikit-learn conventions. 

This EM-based approach compares favorably to scikit-learn. To demonstrate, we generate a dataset of 100k samples in 10 dimensions from 2 overlapping Gaussian ellipses with means of 0 and 1 respectively and standard deviations of 2. It took pomegranate $\sim$0.04s to learn a Gaussian naive Bayes model with 10 iterations of EM, $\sim$0.2s to learn a multivariate Gaussian Bayes classifier with a full covariance matrix with 10 iterations of EM, whereas the scikit-learn label propagation model with a RBF kernel did not converge after $\sim$220s and 1000 iterations, and took $\sim$2s with a knn kernel with 7 neighbors. Both pomegranate models achieved validation accuracies over 0.75, whereas the scikit-learn models did no better than chance.

\vspace{1em}\noindent\textbf{Parallelism:} Another benefit of the use of additive sufficient statistics is that it presents a clear data-parallel way to parallelize model fitting. Simply, one would divide the data into several batches and calculate the sufficient statistics for each batch locally. These sufficient statistics can then be added together back on the main job and all parameters updated accordingly. This is implemented by dividing the data into batches and running \code{summarize} on each of them using separate threads and then running \code{from\_summaries} after all threads finish. Typically, the global interpreter lock (GIL) in Python prevents multiple threads from running in parallel in the same python process. However, since the computationally intensive aspects are written in Cython the GIL can be released, allowing for multiple threads to run at once. On a synthetic data set with 3M samples with 1K dimensions it takes $\sim$65 seconds to train a Gaussian naive Bayes classifier using pomegranate with 1 thread, but only $\sim$17 seconds with 8 threads. For comparison, it takes $\sim$53 seconds to train a Gaussian naive Bayes classifier using scikit-learn. On another synthetic data set with 2M samples and 150 dimensions it takes pomegranate $\sim$470s to learn a Gaussian mixture model with a full covariance matrix with 1 thread, $\sim$135s with 4 threads, $\sim$57s with 16 threads, and $\sim$200s using a GPU. Lastly, we compared the speed at which pomegranate and hmmlearn could train a 10 state dense Gaussian hidden Markov model with diagonal covariance matrices. On a synthetic data set of 100 sequences, each containing 1,000 10 dimensional observations, it took hmmlearn $\sim$25s to run five iterations of Baum-Welch training, while it only took pomegranate $\sim$13s with 1 thread, $\sim$4s with 4 threads, and $\sim$2s with 16 threads.  

\section{Discussion}
pomegranate aims to fill a niche in the Python ecosystem that exists between classic machine learning methods and Bayesian methods by serving as an implementation of flexible probabilistic models. The design choices that were made early on while building pomegranate allowed for a great number of useful features to be added later on without significant effort.

A clear area of improvement in the future is the handling of missing values, because many probabilistic models can intuitively modify the EM algorithm to infer these missing values. For example, when trying to learn a Bayesian network over a dataset with missing values, one can identify the best structure over the incomplete dataset, infer the missing values, and relearn the structure, iterating until convergence. Given the prevalence of missing data in the real world, extending pomegranate to handle missing data efficiently is a priority.

\subsubsection*{Acknowledgments}
We would like to first acknowledge all of the contributors and users of pomegranate, whom without this project would not be possible. We would also like to acknowledge Adam Novak, who wrote the first iteration of the hidden Markov model code. Lastly, we would also like to acknowledge Dr. William Noble for suggestions and guidance during development. This work was partially supported by NSF IGERT grant DGE-1258485.

\bibliography{main}
\bibliographystyle{plainnat}

\end{document}